%% file: ijcai26.tex
\documentclass{article}
\usepackage{ijcai26}

\usepackage{times}
\usepackage{soul}
\usepackage{url}
\usepackage[hidelinks]{hyperref}
\usepackage[utf8]{inputenc}
\usepackage[small]{caption}
\usepackage{graphicx}
\usepackage{amsmath}
\usepackage{amsthm}
\usepackage{booktabs}
\usepackage{algorithm}
\usepackage[switch]{lineno}
\usepackage{algpseudocode}
\usepackage{amssymb}
\usepackage{makecell}
\usepackage{multirow}
\usepackage{subcaption}
\usepackage{tikz}
\usepackage{xspace}

\newtheorem{theorem}{Theorem}
\newtheorem{assumption}{Assumption}

\newcommand{\proposal}{HARMONY\xspace}

\title{\proposal: Bridging the Personalization-Generalization Gap by Mitigating Representation Skew in Heterogeneous Split Federated Learning}

\author{
Jiseok Youn$^1$
\and
You Rim Choi$^2$\and
Goodsol Lee$^1$\and\\
Sangtae Ha$^3$\and
Hyung-Sin Kim$^{2*}$\And
Saewoong Bahk$^1$\thanks{Corresponding authors.}\\
\affiliations
$^1$Department of ECE and INMC, Seoul National University, Seoul, South Korea\\
$^2$Graduate School of Data Science, Seoul National University, Seoul, South Korea\\
$^3$Department of Computer Science, University of Colorado Boulder, CO, USA\\
\emails
jsyoun@netlab.snu.ac.kr, yrchoi@snu.ac.kr, gslee2@netlab.snu.ac.kr, \\ sangtae.ha@colorado.edu, \{hyungkim, sbahk\}@snu.ac.kr
}
\begin{document}

\maketitle

\begin{abstract}
\input{./Section/0_abstract}
\end{abstract}

\section{Introduction}
\label{sec:introduction}
\input{./Section_yr/1_introduction}

\section{Related Work}
\input{./Section/2_related_work}

\section{Proposed Method}
% \input{./Section/3_method}
\input{./Section_yr/3_method}

\section{Convergence Analysis}
\input{./Section/4_analysis}

\section{Evaluation}
\label{sec:evaluation}
\input{./Section/5_evaluation}

\section{Discussion and Future Work}
\label{sec:discussion}
\input{./Section/6_discussion}

\section{Conclusion}
\label{sec:conclusion}

\input{./Section/7_conclusion}

\appendix

% \section*{Ethical Statement}

% There are no ethical issues.

\iffalse
\section*{Acknowledgments}
\fi

%% The file named.bst is a bibliography style file for BibTeX 0.99c
\bibliographystyle{named}
\bibliography{ijcai26}

\end{document}

%% file: Section/0_abstract.tex
Mobile devices face diverse resource constraints and non-IID data class distributions, requiring fast on-device inference for local in-distribution (ID) classes and on-demand remote support for client-specific out-of-distribution (OOD) classes. Hybrid split federated learning (Hybrid SFL) couples personalized client-side front ends (supporting early exit) with a generalized server-side backend for fallback inference, balancing accuracy and cost. However, under client architectural heterogeneity, the existing hybrid SFL suffers from representation skew, where features from customized extractors fail to align in the shared space, leading to a sharp degradation in the server model responsible for OOD prediction. We propose \proposal, the first hybrid SFL framework to support heterogeneous client architectures. \proposal modifies meta-learning to simulate diverse extractors across parameters and architectures, and to learn to personalize. To mitigate representation skew, \proposal conducts server-side contrastive learning to align extracted features, neither sacrificing clients' personalization nor sharing raw labels. Compared to the state of the art across multiple datasets and model families, \proposal improves test accuracy by up to 43.0\%/28.3\% without/with OOD, respectively, while maintaining acceptable latency.

%% file: Section_yr/1_introduction.tex
Federated learning (FL) enables privacy-preserving collaborative training across distributed mobile devices without centralizing sensitive user data, a capability increasingly demanded as personal devices proliferate~\cite{McmahanMRHy:17,WuWWLSG:24,YanLHLB:25,YeFDCT:23}.
However, practical deployment in mobile ecosystems remains challenging due to two fundamental sources of heterogeneity.
First, \textit{resource heterogeneity} arises from diverse compute, memory, and energy budgets, often forcing clients to run undersized or shallow on-device models.
Second, \textit{data heterogeneity} (non-IID label skew) can severely degrade generalization, particularly when a client receives queries from classes that are absent or extremely rare in its local training data.
We refer to such locally unseen or rare classes as \emph{client-specific OOD} and use \emph{OOD} to denote this setting throughout, although these classes may still belong to the global label space~\cite{LiSTS:20,YuHWWZ:23,ChenV:24,HanKCNMCB:24}.

As a promising solution, hybrid split federated learning (Hybrid SFL) has recently emerged by partitioning the model into a personalized on-device model specialized for local in-distribution (ID) classes and a deeper, higher-capacity server model for broader scenarios, including OOD cases.
This split design enables clients to terminate inference locally for confident ID inputs while offloading uncertain (often OOD) cases for server-side fallback, thereby balancing accuracy with computation, communication, and latency.
A representative example is SplitGP~\cite{HanKCNMCB:24}, which jointly optimizes the client exit and server fallback losses and aggregates parameters across rounds.

\begin{figure}[t] 
\centering
    \includegraphics[width=\linewidth]{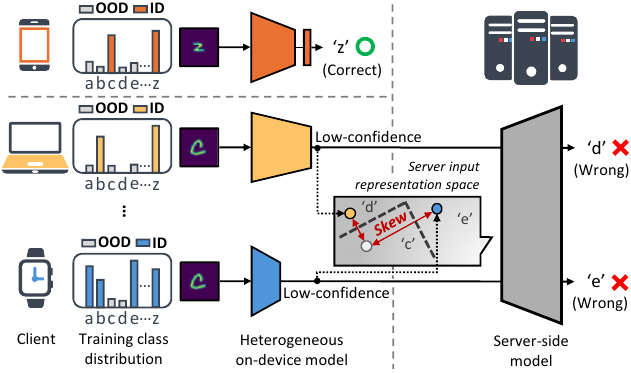} 
\caption{Representation skew in hybrid SFL under architectural and parameter heterogeneity.
Heterogeneous, personalized client front ends misalign embeddings at the split point, degrading server-side fallback.}
\label{fig:motivation}
\end{figure}

Despite this promise, most existing hybrid SFL methods implicitly assume \textit{architectural homogeneity} across clients, i.e., a consistent split point where clients offload embeddings to the server.
In practice, heterogeneous devices instantiate different on-device front ends (e.g., different prefix depths) under diverse resource budgets, and these front ends are further personalized to local label skew.
We find that this combination induces \textit{representation skew}: offloaded embeddings become misaligned across clients, scattering same-class features and spuriously mixing different classes in the server input space (Figure~\ref{fig:motivation}), especially under strong personalization.

\begin{figure}[t]
  \centering
  \begin{subfigure}[t]{0.49\linewidth}
    \centering
    \includegraphics[width=\linewidth]{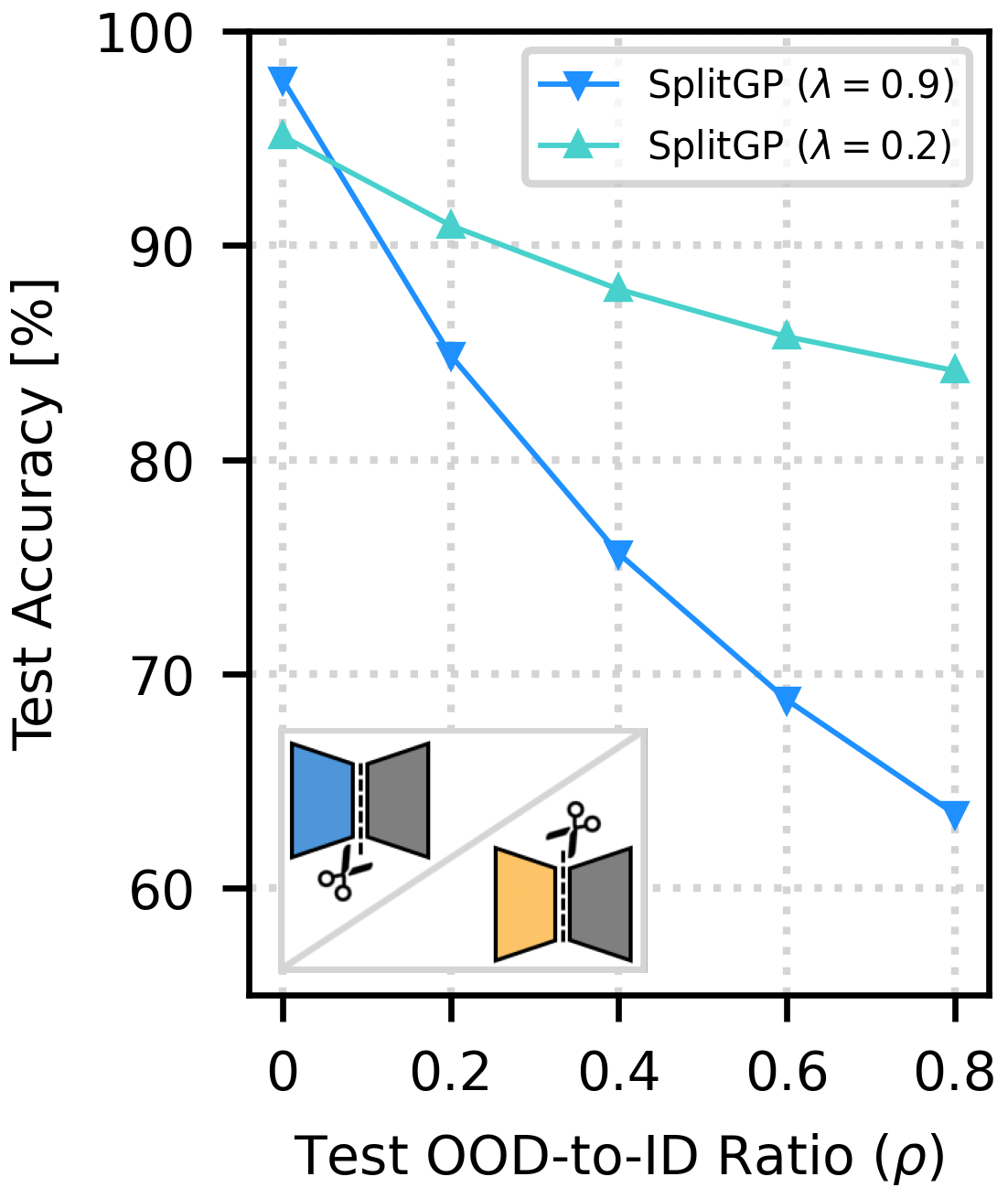}
    \caption{Homogeneous}\label{fig:observation_a}
  \end{subfigure}
  \begin{subfigure}[t]{0.49\linewidth}
    \centering
    \includegraphics[width=\linewidth]{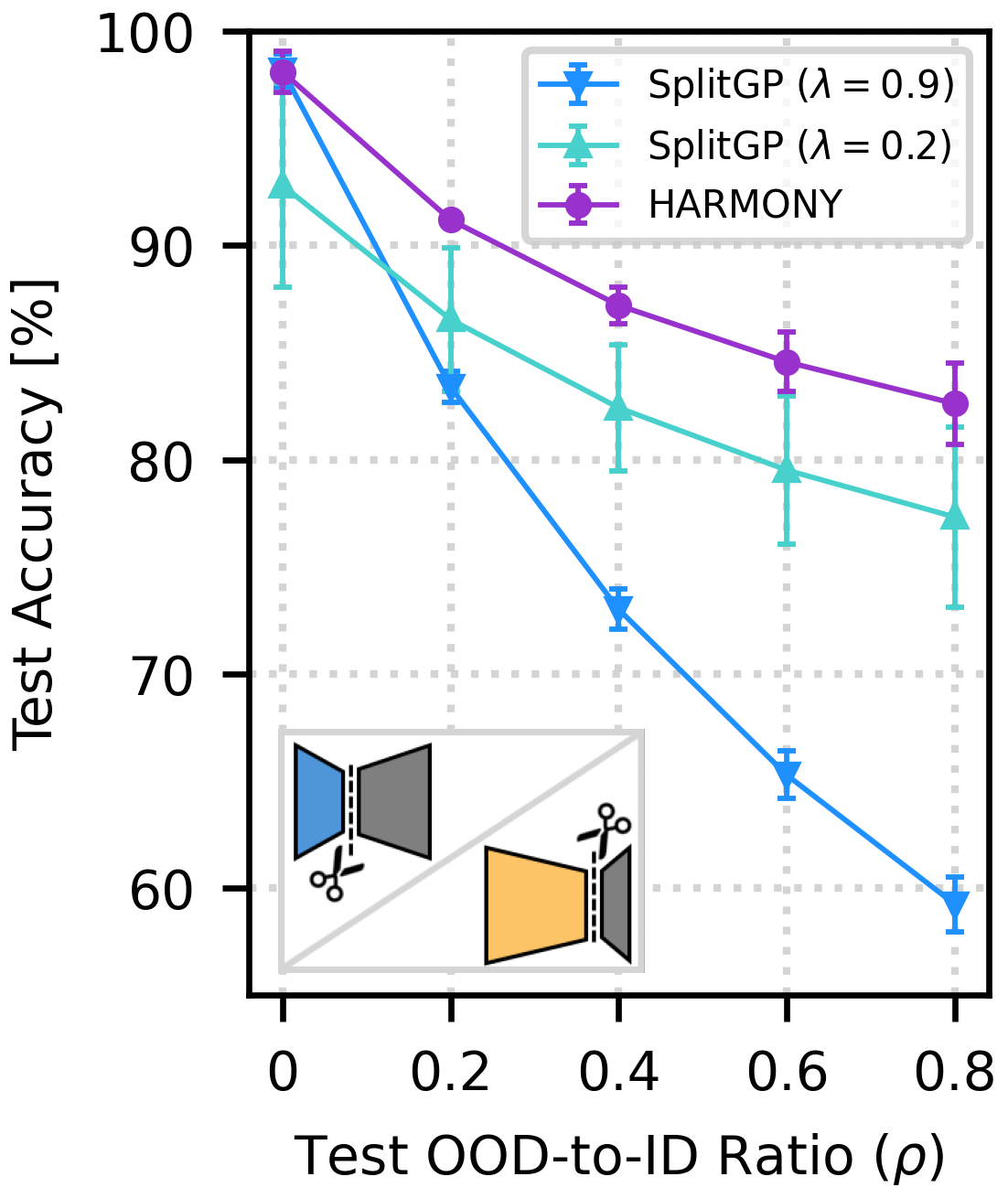}
    \caption{Heterogeneous}\label{fig:observation_b}
  \end{subfigure}
  \caption{SplitGP test accuracy on Fashion-MNIST averaged over 50 clients under varying OOD-to-ID ratio $\rho$ for (a) homogeneous vs.\ (b) heterogeneous client architectures. $\lambda$ denotes personalization intensity. Panel (a) is adapted from Table~V of the SplitGP paper.}
  \label{fig:observation}
\end{figure}

To quantify its impact, we evaluate SplitGP on Fashion-MNIST with a non-IID partition over 50 clients, varying the test OOD-to-ID ratio $\rho=(\#\text{OOD})/(\#\text{ID})$, where each client’s training set contains at most two ID classes and the remaining classes act as OOD for that client.
Smaller $\rho$ favors on-device early exit, whereas larger $\rho$ induces predominantly server-aided inference.
Using VGG-11 as the backbone family, Figure~\ref{fig:observation_a} isolates \emph{parameter heterogeneity} with homogeneous client architectures, while Figure~\ref{fig:observation_b} introduces \emph{architectural heterogeneity} by assigning different split depths to clients.
With homogeneous architectures, reducing personalization from $\lambda{=}0.9$ to $\lambda{=}0.2$ substantially alleviates server degradation at $\rho{=}0.8$ (63.46\% $\to$ 84.15\%).
Under architectural heterogeneity, however, the same adjustment yields limited recovery at $\rho{=}0.8$ (59.25\% $\to$ 77.32\%) and also reduces on-device accuracy at $\rho{=}0$ (98.18\% $\to$ 92.80\%).
Overall, weakening personalization alone is insufficient in heterogeneous hybrid SFL; mitigating representation skew requires explicit cross-client alignment without sacrificing on-device performance.

%\iffalse
%\fi

To address this, we propose \proposal, a heterogeneity-aware hybrid SFL framework that mitigates \textit{representation skew} without sacrificing on-device personalization (82.61\% at $\rho=\:$0.8, 98.10\% at $\rho=\:$0 in Figure~\ref{fig:observation_b}).
\proposal enforces class-aware consistency in the shared server representation space through supervised contrastive alignment, pulling same-class features together across clients while separating different classes.

However, realizing supervised contrastive alignment in hybrid SFL is non-trivial because the split setting imposes three practical requirements.
First, training should remain parallel across clients with a server back end, while still exposing the server to a \emph{diverse neighborhood} of personalized client feature spaces for mitigating representation skew.
We achieve this via \textbf{Meta-Adaptation for Extractor Diversity} (Section~\ref{sec:method_meta}), which promotes fast personalization while providing diverse feature views for server-side alignment.
Second, clients may offload representations at different depths due to device budgets, so alignment must be robust to varying split interfaces.
We address this with \textbf{Stochastic Early Feature Extraction} (Section~\ref{sec:method_stochastic}), which trains the server to handle split-depth variation.
Third, privacy-sensitive labels should remain on-device, yet feature alignment (or task learning) requires class-aware signals.
We therefore use \textbf{Privacy-Preserving Supervision} (Section~\ref{sec:method_privacy}) to enable semantic alignment and fallback training without exposing raw labels.

To summarize, our contributions are:
\begin{itemize}
    \item We identify representation skew as a key bottleneck in hybrid SFL under architectural and parameter heterogeneity, which degrades server-side fallback despite strong on-device personalization.
    \item We propose \proposal (\textbf{H}eterogeneity-\textbf{A}ware \textbf{R}epresentation alignment and \textbf{M}eta-learning for \textbf{O}n-device perso\textbf{N}alization s\textbf{Y}nergy), the first hybrid SFL framework that supports heterogeneous client architectures while mitigating skew via contrastive alignment and meta-adaptation.
    \item We provide a convergence analysis showing that the joint client--server training dynamics asymptotically approach a stationary point under standard assumptions in distributed collaborative learning.
    \item Extensive experiments across multiple datasets and model families demonstrate substantial gains under model heterogeneity, including 43.0\%/28.3\% test accuracy improvement without/with OOD at an acceptable inference-time cost.
\end{itemize}

%% file: Section/2_related_work.tex
\noindent
\textbf{Split Federated Learning}
Split learning~\cite{GuptaR:18} partitions a network across clients and a server to reduce on-device compute and avoid raw-data sharing.
Split federated learning (SFL) combines split training with FL-style parallelism and aggregation across clients (e.g., SplitFed~\cite{ThapaCCS:22}, AdaptSFL~\cite{LinQWCK:25}, FSL-SAGE~\cite{NairLJTSL:25}).
Building on SFL, \emph{hybrid} SFL targets deployment-time efficiency by coupling a personalized client-side front end (enabling early exit) with a generalized server-side back end for fallback inference.
SplitGP~\cite{HanKCNMCB:24} explicitly pursues this personalization--generalization trade-off via a client--server split model with uncertainty-aware fallback, and related work~\cite{Kim2025Personalized} further explores robustness of early-exit SFL (e.g., under label shifts).
However, existing hybrid SFL typically assumes architecturally homogeneous (or parameter-compatible) client front ends; architectural heterogeneity exacerbates representation skew in the shared server space, where same-class features from diverse client backbones fail to align, undermining fallback accuracy.

% \textbf{Split Federated Learning (SFL).} SplitFed~\cite{ThapaCCS:22} pioneered the combination of FL with split learning~\cite{GuptaR:18} to achieve the strengths of both approaches. Building on this foundation, several works have focused on enhancing training efficiency and inference accuracy. GAS~\cite{YangL:25} tackles the straggler problem through asynchronous SFL with generative activations, while RingSFL \cite{ShenCWLXLAS:24} mitigates straggler-induced delays (i.e., slow clients) via a ring-topology architecture. AdaptSFL~\cite{LinQWCL:24} reduces training time by jointly optimizing model splitting and aggregation based on theoretical analysis. Hybrid approaches, such as SplitGP~\cite{HanKCNMCB:24} and PHSFL~\cite{PervejM:24}, balance personalization and generalization through weighted aggregation and two-stage training, respectively. However, these methods typically assume homogeneous model architectures, limiting their applicability in real-world heterogeneous environments.

\noindent
\textbf{Model-Heterogeneous FL}
Model-heterogeneous FL relaxes architectural compatibility by replacing parameter averaging with cross-model knowledge transfer or representation coupling.
A common approach is distillation: FedMD~\cite{li2019fedmd} transfers predictive knowledge via shared unlabeled data, while FedDF~\cite{lin2020ensemble} performs server-side ensemble distillation for robust fusion.
Beyond logits, representation-level coupling is also used: FedProto~\cite{tan2022fedproto} exchanges class prototypes to encourage semantic consistency, and FCCL~\cite{huang2022learn} leverages unlabeled public data with a cross-correlation objective to learn heterogeneity-robust representations. FedSKD~\cite{WengCZ:25} introduces round-robin model circulation and multi-dimensional similarity distillation to remove dependence on a centralized server. FedFD~\cite{LiWXWQDL:25} implements feature distillation by architecture-wise feature aggregation and orthogonality-aided projection layers.
However, these methods are not tailored to hybrid SFL constraints (early exit and server fallback) nor to customization-induced representation skew at the split interface; mitigating this skew requires explicit representation alignment for reliable fallback inference.

% \noindent
% \textbf{Heterogeneous Federated Learning.} Classical FL addresses device heterogeneity, including differences in computing capability, data distribution, and model architecture, through various algorithmic innovations. FedProx~\cite{LiKZSTS:20} introduces a proximal term for partial updates under resource constraints, while meta-learning approaches ~\cite{FallahMO:20,LiHBS:21,OhKY:22,YuanPABC:24,TamirisaXBZAS:24} enable rapid adaptation to local distributions. For SFL specifically, ParallelSFL~\cite{LiaoXXYHQ:24} partitions workers by utility, MultiSFL \cite{XiaHYLLXC:25} uses knowledge replay to address statistical heterogeneity, and recent work~\cite{HanHTTL:24} provides convergence guarantees under data heterogeneity. 
% % Nonetheless, existing methods often assume monolithic models or homogeneous SFL architectures, leaving architectural heterogeneity in SFL largely unaddressed.
% Nonetheless, existing methods often assume monolithic or homogeneous-architecture models, leaving model heterogeneity in SFL largely unaddressed.

%% file: Section_yr/3_method.tex
%%%%%%%%%%%%
\newcommand{\bcircled}[2][black]{%
  \tikz[baseline=(char.base)]{%
    \node[shape=circle, fill=#1, text=white, draw=none,
          inner sep=1pt,
          minimum size=0.8em,
          font=\sffamily\small\bfseries] (char) {#2};}}
%%%%%%%%%%%% 

\begin{figure}[t]
\centerline{\includegraphics[width=0.5\textwidth]{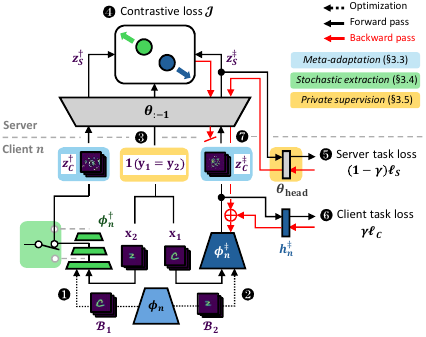}}
% \caption{Overview of \proposal.}
\caption{Overview of local training process by \proposal, with highlights on our solutions mentioned in Chapter~\ref{sec:introduction}.}
\label{fig:ours_overview}
\end{figure}

%%%%%%%%%%%%
In this section, we present \proposal, a hybrid SFL framework that mitigates \emph{representation skew}---the misalignment between highly personalized, potentially heterogeneous client-side extractors and the shared server back end---to preserve strong on-device personalization while maintaining reliable server-side fallback generalization.
% \proposal achieves this by exposing the server to diverse personalized feature spaces and aligning them via privacy-preserving pairwise supervision.
\proposal achieves this by exposing the server model to diverse feature (client output) spaces and aligning them via privacy-preserving pairwise supervision. 

Algorithms of \proposal can be found in appendix A.
%%%%%%%%%%%%

%%%%%%%%%%%%
\subsection{Problem Setup and Preliminaries}
\label{sec:method_prelim}

We consider an SFL scenario with $N$ clients, a main server, and a fed server.
Each \emph{client} $n\,(=1,\dots,N)$ holds a local dataset $\mathcal{D}_n$ and a client-side model consisting of (i) a resource-adaptive front end $\phi_n$, instantiated as a prefix of a common multi-exit backbone template with a client-specific exit/split point, and (ii) a classifier head $h_n$ for on-device inference (local exit).
The \emph{main server} hosts a high-capacity back-end model $\theta$, including the final head, for fallback inference.
% when needed.
% The \emph{fed server} is used only for training to aggregate client-side models under mutual trust of model privacy~\cite{ThapaCCS:22}.
During SFL, the \emph{fed server} aggregates $\{\phi_n\}$ and $\{h_n\}$ under mutual trust of model privacy~\cite{ThapaCCS:22}.

For each round $r=0,\dots,R{-}1$, the system selects participants $\mathbb{P}_r$ and the main server duplicates $\theta^{r}$ to $\{\theta_{n}^{r,t=0}\}_{n\in\mathbb{P}_r}$, enabling $T_n$ steps of parallel split optimization for each client--server pair.
Following the SplitGP protocol~\cite{HanKCNMCB:24}, at each local step $t\,(=0,\dots,T_n{-}1)$ for client $n$, the client and server optimize a weighted combination of the on-device and server task losses:
\begin{equation}
\begin{aligned}
\ell_{C}&\gets\mathcal{L}\bigl((h_{n}^{r,t}\circ\phi_{n}^{r,t})(\mathbf{x}),\mathbf{y}\bigr), \\
\ell_{S}&\gets\mathcal{L}\bigl((\theta_{n}^{r,t}\circ\phi_{n}^{r,t})(\mathbf{x}),\mathbf{y}\bigr),
\end{aligned}
\label{eq:splitgp_loss}
\end{equation}
where $(\mathbf{x},\mathbf{y})\sim\mathcal{D}_n$ and $\mathcal{L}(\cdot)$ is the downstream-task loss.
Parameters are updated to minimize $\gamma\,\ell_C+(1-\gamma)\,\ell_S$, where $\gamma\in[0,1]$ balances client personalization and server generalization.
After local optimization, the fed server aggregates client-side models with coefficient $\lambda$, while the main server aggregates $\theta$ via FedAvg~\cite{McmahanMRHy:17}.
% We use $\zeta_n=\frac{|\mathcal{D}_n|}{\sum_i|\mathcal{D}_i|}$ to weight each client in aggregation:
$\zeta_n=\frac{|\mathcal{D}_n|}{\sum_i|\mathcal{D}_i|}$ considers local data size imbalance.
Because clients may instantiate $\phi_n$ (and the exit head $h_n$) with different prefix depths, the fed server performs depth-aware partial aggregation, averaging only the blocks that exist in each client model (e.g., HeteroFL~\cite{DiaoDT:21}).
\begin{equation}\label{eq:splitgp_aggregation}
\begin{aligned}
\phi_{n}^{r+1} &= \lambda\,\phi_{n}^{r,T_{n}}+(1-\lambda)\sum_{i}\zeta_{i}\,\phi_{i}^{r,T_{i}},\\
h_{n}^{r+1} &= \lambda\,h_{n}^{r,T_{n}}+(1-\lambda)\sum_{i}\zeta_{i}\,h_{i}^{r,T_{i}},\\
\theta^{r+1} &= \sum_{i}\zeta_{i}\,\theta_{i}^{r,T_{i}}.
\end{aligned}
\end{equation}

At inference, client $n$ exits on-device if $E(\mathrm{Softmax}((h_n\circ\phi_n)(\mathbf{x})))<e_n$; otherwise, it offloads the client embedding $\mathbf{z}_{C}=\phi_{n}(\mathbf{x})$ to the main server and receives $\theta(\mathbf{z}_{C})$.
Here $E(\cdot)$ denotes Shannon entropy~\cite{Shannon:48} and $e_{n}$ is the threshold.
We follow this inference protocol throughout.

%%%%%%%%%%%%
\subsection{Overview and Design Rationale}
\label{sec:method_overview}

Hybrid SFL must satisfy three practical requirements:
(i) split training should proceed in parallel across clients with a server back end,
(ii) clients may have different split depths due to device budgets, and
(iii) privacy-sensitive labels should remain on-device.
Meanwhile, mitigating representation skew requires the server to receive \emph{semantic} (class-aware) signals that are robust to heterogeneous client front ends.

Figure~\ref{fig:ours_overview} illustrates one local step of \proposal.
Each client simulates two customized extractors and generates two feature views $(\mathbf{z}_C^{\dagger},\mathbf{z}_C^{\ddagger})$ sent to the main server.
Since representation skew is \emph{class-agnostic}, the server must receive class-aware supervision to organize heterogeneous client features by class \emph{without uploading raw labels}.
Accordingly, the server performs two complementary updates:
(a) label-private task training to improve fallback accuracy, and
(b) contrastive semantic alignment to enforce class-aware structure in the shared server representation space.
Importantly, task gradients are propagated across the split to update both client and server modules, whereas contrastive gradients are confined to the server to avoid interfering with on-device personalization.
We next describe how \proposal realizes this workflow via three modules.

%%%%%%%%%%%%
\subsection{Meta-Adaptation for Extractor Diversity}
\label{sec:method_meta}

\paragraph{Purpose and meta-learning principle.}
This module serves two goals.
First, it trains each client extractor to become \emph{rapidly personalizable}: starting from the current parameters $(\phi_n^{r,t},h_n^{r,t})$, a few SGD steps on a small local batch should yield an effective personalized model. As a result, the need for personalization via tuning the aggregation is eliminated; therefore, HARMONY sets $\lambda=0$.
Second, it exposes the server to a \emph{diverse neighborhood} of personalized feature spaces induced by such fast adaptation, which is crucial for mitigating personalization-induced representation skew under heterogeneous clients.
To achieve both, \proposal follows the core idea of MAML~\cite{FinnAL:17}: an \emph{inner loop} performs temporary few-step adaptation into a batch, and an \emph{outer loop} updates the pre-adaptation parameters so that future few-step personalization becomes more effective.

\paragraph{Two-branch cross-batch adaptation.}
As illustrated in Figure~\ref{fig:ours_overview} (\bcircled{1}--\bcircled{2}), 
at each local step $t$ in round $r$, client $n$ samples two mini-batches
$\mathcal{B}_{1}=(\mathbf{x}_{1},\mathbf{y}_{1})$ and $\mathcal{B}_{2}=(\mathbf{x}_{2},\mathbf{y}_{2})$ from $\mathcal{D}_n$.
Starting from the current client parameters $(\phi_n^{r,t},h_n^{r,t})$ (with $(\phi_n^{r,0},h_n^{r,0})\leftarrow(\phi^{r},h^{r})$ from the fed server), the client runs two separate few-step inner-loop adaptations to obtain two temporary models personalized into the corresponding batch:
\begin{equation}
\begin{aligned}
(\phi_n^{\dagger},h_n^{\dagger}) \leftarrow \textsc{Adapt}\bigl((\phi_n^{r,t},h_n^{r,t});\mathcal{B}_1\bigr),\\
(\phi_n^{\ddagger},h_n^{\ddagger}) \leftarrow \textsc{Adapt}\bigl((\phi_n^{r,t},h_n^{r,t});\mathcal{B}_2\bigr),
\end{aligned}
\end{equation}
where $\textsc{Adapt}(\cdot)$ denotes a small number of SGD steps.
We then use these adapted models in a \emph{cross-batch} manner to construct two complementary feature views:
\begin{equation}\label{eq:views}
\mathbf{z}_C^{\dagger}\; \leftarrow\; \{{\phi}_{n}^{\dagger}\}_{:K+1}(\mathbf{x}_{2}),
\qquad
\mathbf{z}_C^{\ddagger}\; \leftarrow\; \phi_{n}^{\ddagger}(\mathbf{x}_1).
\end{equation}
where branch $\dagger$ extracts an intermediate representation with a stochastically sampled exit/split depth $K$ (Section~\ref{sec:method_stochastic}), while branch $\ddagger$ extracts a task-relevant representation.
This cross-batch construction yields two feature streams from different local samples within the same client step, which are later paired to provide semantic signals for server-side contrastive alignment under fast adaptation (Section~\ref{sec:method_privacy}).

\paragraph{Meta-update (outer loop).}
The adapted models $(\phi_n^{\dagger},h_n^{\dagger})$ and $(\phi_n^{\ddagger},h_n^{\ddagger})$ are temporary and used only to generate feature views and their associated learning signals.
The actual optimization target is the pre-adaptation client model $(\phi_n^{r,t},h_n^{r,t})$, updated using the losses computed from the two views (defined in Section~\ref{sec:method_privacy}).
Since the losses are evaluated on inner-loop adapted models, the exact meta-gradient involves differentiating through the adaptation steps, which introduces second-order (Hessian) terms; for efficiency, we use the first-order approximation (FOMAML)~\cite{nichol2018first}.

\paragraph{Personalization after SFL.} At last, client $n$ personalizes:
\begin{equation}
\begin{aligned}
(\phi_n,h_n) \leftarrow \textsc{Adapt}\bigl((\phi^{R},h^{R});\mathcal{D}_{n}\bigr).
\end{aligned}
\end{equation}

%%%%%%%%%%%%
\subsection{Stochastic Early Feature Extraction}
\label{sec:method_stochastic}

\proposal stochastically varies the exit/split depth when forming the branch $\dagger$ view.
This emulates heterogeneous device budgets during training: although clients share a common multi-exit backbone family, they may offload features at different depths, so the server must be robust to \emph{split-interface variation}.
After obtaining the adapted model $\phi_n^{\dagger}$, client $n$ samples an exit index $K$ and offloads the corresponding intermediate activation $\mathbf{z}_C^{\dagger}$ (Eq.~\eqref{eq:views}).
The main server then produces the server-side embedding via \emph{depth-matched routing}:
it keeps the back end $\theta$ as the largest model and feeds $\mathbf{z}_C^{\dagger}$ into the block at the corresponding depth, skipping earlier blocks.
Formally,
\begin{equation}
\mathbf{z}_S^{\dagger} \;\leftarrow\; \theta_{K'+1:-1}(\mathbf{z}_C^{\dagger}), \quad \text{where}\:\theta_{K'}\equiv(\theta\,\circ\,\phi)_{K}
\end{equation}
where $\theta_{K'+1:-1}$ denotes the part of $\theta$ after split depth $K'$ (the prefix $\theta_{:K'+1}$ is bypassed) and before $\theta_{\text{head}}$.
% By sampling $K$ across steps and clients, the server is trained on a mixture of split interfaces, improving robustness to client-specific exit/split points under architectural heterogeneity.

%%%%%%%%%%%%
\subsection{Privacy-Preserving Supervision}
\label{sec:method_privacy}

This module provides the main server with \emph{class-aware} supervision to mitigate representation skew, while keeping raw labels on-device.
As illustrated in Figure~\ref{fig:ours_overview} (\bcircled{3}--\bcircled{7}), \proposal combines
(i) proxy-supervised contrastive alignment from a binary label-matching indicator and
(ii) label-private task training via a U-shaped split protocol.

\paragraph{Proxy-supervised contrastive alignment (\bcircled{3}--\bcircled{4}).}
To inject semantic structure without revealing labels, client $n$ transmits the paired features $(\mathbf{z}_C^{\dagger},\mathbf{z}_C^{\ddagger})$ together with a binary indicator
$\mathrm{I}\triangleq \mathbf{1}(\mathbf{y}_1=\mathbf{y}_2)$.
On the server, we compute pre-head logit embeddings using the server trunk $\theta_{:-1}$ (Figure~\ref{fig:ours_overview}):
$\mathbf{z}_S^{\dagger} \triangleq \theta_{K'+1:-1}(\mathbf{z}_C^{\dagger})$ and
$\mathbf{z}_S^{\ddagger} \triangleq \theta_{\{n(\phi)\}':-1}(\mathbf{z}_C^{\ddagger})$,
where $K'$ and $\{n(\phi)\}'$ denote the depth-matched suffix execution for the stochastically chosen split depth $K$ (Section~\ref{sec:method_stochastic}) and the number of client's layers $n(\phi)$, respectively.
We then apply contrastive semantic alignment (CSA)~\cite{MotiianPAD:17}:
\begin{equation}
\ell_{\text{CSA}}
=
\mathrm{I}\cdot \tfrac{1}{2}\|\mathbf{z}_S^{\dagger}-\mathbf{z}_S^{\ddagger}\|^{2}
+
(1-\mathrm{I})\cdot \tfrac{1}{2}\max\!\bigl(0,\, m-\|\mathbf{z}_S^{\dagger}-\mathbf{z}_S^{\ddagger}\|\bigr)^{2},
\end{equation}
which pulls same-class pairs together and repels different-class pairs, directly counteracting class-agnostic skew in the shared server space.
Importantly, CSA gradients are confined to the server: we stop gradients at $(\mathbf{z}_C^{\dagger},\mathbf{z}_C^{\ddagger})$ and update only $\theta_{:-1}$ to avoid interfering with on-device personalization.

\paragraph{Label-private task training (\bcircled{5}--\bcircled{7}).}
In parallel, \proposal trains the fallback path without uploading labels by using a U-shaped split protocol~\cite{vepakomma2018split}.
Concretely, the server performs the forward pass on $\mathbf{z}_C^{\ddagger}$ and returns $\mathbf{z}_S^{\ddagger}$ to the client; the client evaluates the server task loss $\ell_S$ locally using its private label and sends back only the upstream gradient, enabling the server to update $\theta$ while labels never leave the device.
The client simultaneously optimizes its on-device loss $\ell_C$; the overall task objective follows the hybrid weighting $\gamma\,\ell_C+(1-\gamma)\,\ell_S$ defined in Section~\ref{sec:method_prelim}.

%% file: Section/4_analysis.tex
In this section, we provide the convergence behavior of \proposal. Let $v \triangleq \big(\{\phi_n,h_n\}_{n=1}^N,\theta\big)$ and $p_{n}$ be the participation ratio. We define the multi-exit and global objectives:

\begin{equation}
  F_n(\phi_n,h_n,\theta) \;=\; \gamma\,\ell_{C}(\phi_n,h_n) \;+\; (1-\gamma)\,\ell_{S}(\theta;\phi_n),
\end{equation}
\begin{equation}
  \mathcal{F}(v)
  \;=\; \sum_{n} p_{n}\Big[F_n(\phi_n,h_n,\theta)+ \mathcal{J}\big(\phi_n,\theta\big)\Big].
  \label{eq:objectives}
\end{equation}

We adopt known assumptions for distributed ML related to our components such as meta-learning, contrastive learning, and quantization, explained below:

\begin{assumption}[Smoothness~\cite{WangLLJP:20}]\label{asm:smooth}
$\mathcal{F}$ is $L$-smooth: $\|\nabla \mathcal{F}(x)-\nabla \mathcal{F}(y)\|\le L\|x-y\|,\ \forall x,y$.
\end{assumption}

\begin{assumption}[Client dissimilarity~\cite{LiSZSTS:20}]\label{asm:dissim}
For some $B\ge 1$, $\sum_{n} p_n \|\nabla F_n(v)\|^2 \le B^2 \|\nabla \mathcal{F}(v)\|^2$. 
\end{assumption}

\begin{assumption}[Stochastic gradient noise property~\cite{ReddiCZGRKKM:21}]\label{asm:stoch}
The stochastic gradient noise $\epsilon_{\rm sg}$ satisfies
$\mathbb{E}[\epsilon_{\rm sg}\mid v]=0$ and $\mathbb{E}\left[\|\epsilon_{\rm sg}\|^2\right]\le(\sigma_C^2+\sigma_S^2)$. 
\end{assumption}

\begin{assumption}[Quantization (+ STE) noise property~\cite{ReisizadehMHJP:20}]\label{asm:quant}
The noise $\epsilon_{\rm q}$ by quantizing features before transfer satisfies
$\mathbb{E}[\epsilon_{\rm q}\mid v]=0$ and $\mathbb{E}\left[\|\epsilon_{\rm q}\|^2\right]\le \sigma_{\rm q}^2$.
\end{assumption}

\begin{assumption}[FO-MAML outer-gradient property~\cite{FallahMO:20}]\label{asm:maml}
For inner-loop step-size $\alpha\le k/L$ and an upper bound $G$ on the second moment of the inner-loop gradients, the FO-MAML approximation bias $\epsilon_{\rm maml}$ obeys
$\big\|\mathbb{E}[\epsilon_{\rm maml}\mid v]\big\|\le k'\alpha L G$ and
$\mathbb{E}\left[\|\epsilon_{\rm maml}\|^2\right]\le C_{\rm maml}\cdot\alpha^2$.
\end{assumption}

\begin{assumption}[CSA pair-sampling variance~\cite{SaunshiPAKK:19,ChenGLGCGCXZLLCT:21}]\label{asm:csa}
The contrastive loss $\mathcal{J}$ contributes a zero-mean variance term $\epsilon_{\rm csa}$ on $\theta$-coordinates only:
$\mathbb{E}[\epsilon_{\rm csa}\mid v]=0$, $\mathbb{E}\left[\|\epsilon_{\rm csa}\|^2\right]\le \sigma_{\rm csa}^2$.
\end{assumption}

\begin{assumption}[Step-sizes]\label{asm:step_sizes}
$\alpha\le k/L$, $\beta\le k''/L(B^2+1)$.
\end{assumption}

% The assumptions and following lemmas lead to the convergence theorem:
The assumptions and following lemmas lead to:

\begin{theorem}[Stationary-point convergence]\label{thm:stationary}
Let $\bar T=\sum_n p_n T_n$ and let there be $R$ rounds.
Under Assumptions~\ref{asm:smooth}--\ref{asm:step_sizes},
\begin{equation}
\begin{aligned}
\min_{0\le r<R}\ \mathbb{E}\left[\left\|\nabla\mathcal{F}(v^r)\right\|^2\right]
\ &\le\
\frac{4\big(\mathbb{E}\left[\mathcal{F}(v^0)\right]-\mathcal{F}^{{\setminus}*}\big)}{\beta\,R\bar T}
\;\\ &+\; 20L\beta\left(C_{\rm sg} + C_{\rm q} + C_{\rm csa}\right) \\&+ \left(20L\beta+\frac{\bar{T}}{{L}{\beta}}\right)\alpha^{2}C_{\rm maml},
\label{eq:main-rate}
\end{aligned}
\end{equation}
where $\mathcal{F}^{{\setminus}*}$ is a lower bound on $\mathcal{F}$, and $C_{\rm sg}, C_{\rm q}, C_{\rm maml}, C_{\rm csa}$ are constants related to $\epsilon_{\rm sg}, \epsilon_{\rm q}, \epsilon_{\rm maml}, \epsilon_{\rm csa}$, respectively.
\end{theorem}
The proof can be found in appendix B.

%% file: Section/5_evaluation.tex
\begin{table}[t]
\centering
\small
\setlength{\tabcolsep}{3pt}
\begin{tabular}{llrrrrr}
\toprule
 & & \textbf{8-layer} & \textbf{VGG-11} & \textbf{MV2} & \textbf{RN50} & \textbf{ViT-S} \\
\midrule
\multirow{2}{*}{\(|\phi|\)} & Min. & 18.8 & 76.0 & 0.9 & 84.5 & 3920.3 \\
 & Max. & 387.8 & 962.3 & 239.4 & 884.8 & 11018.1 \\
\multirow{2}{*}{\(|h|\)} & Min. & 5.8 & 5.1 & 3.3 & 51.4 & 39.3 \\
 & Max. & 23.1 & 10.3 & 6.5 & 102.6 & 39.3 \\
\multirow{2}{*}{\(|\theta|\)} & Min. & 3480.3 & 8268.8 & 2112.6 & 23033.0 & 10686.1 \\
 & Max. & 3849.4 & 9155.1 & 2351.0 & 23833.3 & 17783.9 \\
\bottomrule
\end{tabular}
\caption{Minimum/Maximum number of model parameters $[{\times}10^3]$ by diverse model split points across clients.}
\label{table:model_numels}
\end{table}

\begin{table*}[t]
\centering
\small
\setlength{\tabcolsep}{3pt}
\begin{tabular}{llccccc}
\toprule
\textbf{Scheme} & \textbf{Metric} & $\rho{=}0$ & 0.2 & 0.4 & 0.6 & 0.8 \\
\midrule

\multirow{2}{*}{SplitFed}
& Accuracy & 47.93 $\pm$ 1.00 & 46.83 $\pm$ 0.89 & 46.04 $\pm$ 1.07 & 45.46 $\pm$ 1.40 & 45.03 $\pm$ 1.66 \\
& Latency  & 104.80 $\pm$ 4.21 & 125.76 $\pm$ 5.06 & 146.72 $\pm$ 5.90 & 167.68 $\pm$ 6.78 & 188.63 $\pm$ 7.62 \\
\midrule

\multirow{2}{*}{Multi-exit SFL}
& Accuracy & 67.13 $\pm$ 4.04 & 64.69 $\pm$ 4.99 & 63.06 $\pm$ 5.95 & 61.90 $\pm$ 6.79 & 61.01 $\pm$ 7.35 \\
& Latency  & 41.57 $\pm$ 15.10 & 48.57 $\pm$ 19.00 & 54.92 $\pm$ 20.15 & 60.38 $\pm$ 23.14 & 66.54 $\pm$ 34.70 \\
\midrule

\multirow{2}{*}{SplitGP ($\lambda{=}0.9$)}
& Accuracy & 93.00 $\pm$ 0.81 & 78.12 $\pm$ 0.54 & 67.69 $\pm$ 0.21 & 59.96 $\pm$ 0.10 & 54.00 $\pm$ 0.25 \\
& Latency  & 26.50 $\pm$ 1.11 & 38.59 $\pm$ 4.05 & 52.99 $\pm$ 9.65 & 70.71 $\pm$ 6.81 & 86.75 $\pm$ 9.07 \\
\midrule

\multirow{2}{*}{SplitGP ($\lambda{=}0.2$)}
& Accuracy & 88.94 $\pm$ 1.57 & 80.87 $\pm$ 1.06 & 75.13 $\pm$ 1.57 & 70.95 $\pm$ 2.21 & 67.71 $\pm$ 2.74 \\
& Latency  & 20.74 $\pm$ 4.55 & 25.78 $\pm$ 7.40 & 32.66 $\pm$ 10.42 & 40.85 $\pm$ 14.17 & 50.28 $\pm$ 16.89 \\
\midrule

\multirow{2}{*}{SplitGP ($\lambda{=}0.2$) + FT}
& Accuracy & \textbf{93.56 $\pm$ 0.84} & 81.49 $\pm$ 0.49 & 73.66 $\pm$ 0.73 & 68.36 $\pm$ 1.50 & 64.46 $\pm$ 2.13 \\
& Latency  & 20.01 $\pm$ 1.78 & 31.27 $\pm$ 6.54 & 50.73 $\pm$ 6.87 & 72.93 $\pm$ 11.61 & 94.90 $\pm$ 13.02 \\
\midrule

\multirow{2}{*}{\textbf{HARMONY}}
& Accuracy & 92.34 $\pm$ 0.98 & \textbf{83.77 $\pm$ 0.50} & \textbf{78.07 $\pm$ 1.44} 
& \textbf{74.23 $\pm$ 2.31} & \textbf{71.45 $\pm$ 2.82} \\
& Latency  & 21.39 $\pm$ 3.55 & 31.10 $\pm$ 0.95 & 48.23 $\pm$ 3.15 & 68.30 $\pm$ 3.54 & 89.58 $\pm$ 7.13 \\
\bottomrule
\end{tabular}
\caption{Comparison of test accuracy [\%] and latency [${\times}10^{8}$] under varying OOD-to-ID proportion $\rho$, in CIFAR-10 / $N=50$ / Shard-2 / VGG-11 settings. $\lambda$ stands for the personalization intensity by SplitGP. `FT' means on-device fine-tuning.}
\label{table:main_results1}
\end{table*}

\begin{table}[t]
\centering
\small
\setlength{\tabcolsep}{2pt}
\begin{tabular}{llcc}
\toprule
\textbf{Scheme} & \textbf{Metric} & $\rho{=}0$ & 0.49 \\
\midrule

\multirow{2}{*}{SplitFed}
& Accuracy & 6.01 $\pm$ 4.04 & 4.42 $\pm$ 2.88 \\
& Latency  & 1.57 $\pm$ 0.10 & 2.34 $\pm$ 0.14 \\
\midrule

\multirow{2}{*}{Multi-exit SFL}
& Accuracy & 25.93 $\pm$ 2.99 & 18.18 $\pm$ 2.01 \\
& Latency  & 0.52 $\pm$ 0.15 & 0.74 $\pm$ 0.31 \\
\midrule

\multirow{2}{*}{SplitGP ($\lambda{=}0.9$)}
& Accuracy & 82.73 $\pm$ 10.39 & 55.57 $\pm$ 6.96 \\
& Latency  & 0.21 $\pm$ 0.12 & 0.31 $\pm$ 0.18 \\
\midrule

\multirow{2}{*}{SplitGP ($\lambda{=}0.2$)}
& Accuracy & 41.63 $\pm$ 22.77 & 28.56 $\pm$ 14.80 \\
& Latency  & 0.28 $\pm$ 0.13 & 0.40 $\pm$ 0.17 \\
\midrule

\multirow{2}{*}{SplitGP ($\lambda{=}0.2$) + FT}
& Accuracy & 67.62 $\pm$ 27.05 & 45.80 $\pm$ 17.78 \\
& Latency  & 0.21 $\pm$ 0.12 & 0.32 $\pm$ 0.19 \\
\midrule

\multirow{2}{*}{\textbf{HARMONY}}
& Accuracy & \textbf{84.62 $\pm$ 4.32} & \textbf{56.83 $\pm$ 2.89} \\
& Latency  & 0.20 $\pm$ 0.13 & 0.31 $\pm$ 0.20 \\
\bottomrule
\end{tabular}
\caption{Comparison of test accuracy [\%] and latency [${\times}10^{8}$] in CIFAR-100 / $N=30$ / Shard-2 / MobileNet-V2 settings.}
\label{table:main_results2}
\end{table}

\textbf{Datasets and Models.} 
% We evaluate on standard benchmarks: Fashion-MNIST~(FMNIST)~\cite{XiaoRV:17}, CIFAR-10/100~\cite{Krizhevsky:09}, CINIC-10 \cite{DarlowCAS:18}, and Tiny-ImageNet~\cite{LeY:15}. We employ various models: an 8-layer CNN (5 convolutional + 3 fully connected layers), VGG-11 (with BN), MobileNet-V2, ResNet-50, and ViT-S. For architectural heterogeneity, we distribute models with varying split points across clients as shown in Table~\ref{table:model_numels}.
We evaluate on widely used vision benchmarks: Fashion-MNIST (FMNIST)~\cite{XiaoRV:17}, CIFAR-10/100~\cite{Krizhevsky:09}, CINIC-10~\cite{DarlowCAS:18}, and Tiny-ImageNet~\cite{LeY:15}.
We employ diverse backbones, including an 8-layer CNN (5 convolutional + 3 fully connected layers), VGG-11 (with BN), MobileNet-V2, ResNet-50, and ViT-S.
To instantiate model heterogeneity, we assign clients different split points (heterogeneous split depths) as summarized in Table~\ref{table:model_numels}.

\noindent
\textbf{Implementation Details.} 
% We deploy $N\in\{30,50, 100\}$ clients depending on experimental complexity. For non-IID data distribution, we randomly allocate \{2, 3\} out of 100 label-sorted data shards per client (`Shard-\{2, 3\}'), or adopt $\text{Dirichlet}(\alpha_{D})$ distribution with concentration parameter $\alpha_{D}$. To simulate data drift, we inject OOD-label shards into local test data with proportion $\rho$ defined in ~\eqref{eq:rho_def}, or modify the Dirichlet distribution following~\cite{JiangL:23}~(`Out-of-Client').
We deploy $N\in\{30,50,100\}$ clients depending on experimental complexity. 
For non-IID data distribution, we either randomly allocate \{2,3\} of 100 label-sorted shards per client (\texttt{Shard-\{2,3\}}), where a shard is an equal-sized chunk after sorting by label, or use a Dirichlet split with parameter $\alpha_D$.
To simulate data drift, we inject locally unseen-label (client-specific OOD) test samples with proportion $\rho$ in~\eqref{eq:rho_def}, or modify the Dirichlet split following~\cite{JiangL:23} (\texttt{Out-of-Client}).

\begin{equation}\label{eq:rho_def}
\rho=\frac{\text{\# of OOD test samples}}{\text{\# of ID test samples}}
\end{equation}

% We conduct $R{\,\in\,}\{50, 120, 400, 800, 1000\}$ maximum SFL rounds where ${\forall}r{\:\:}|\mathbb{P}_{r}|=N$ unless stated otherwise. Each client performs \{1, 5\} local epoch(s) per round with $\gamma=0.5$ and cross-entropy loss $\mathcal{L}(\,\cdot\,)$. For \proposal, we use SGD with learning rate $\beta{\;\in\;}\{0.025, 0.05\}$, momentum 0.9, and weight-decay \{0, 1, 5\}${\,\times\,}10^{-4}$. Meta-learning employs $\alpha{\,\in\,}\{0.01, 0.05\}$ and $M\,{\in}\,\{3, 5, 10\}$ inner steps.
% We use batch size \{50, 64, 128, 256, 512\} in training. At test, we target batch size 1. If required, each client $n$ selects offloading sensitivity $e_{n}$ maximizing test accuracy (e.g., $e_{n}\in\{0.05, 0.1, 0.2, 0.4, 0.8, 1.2, 1.6, 2.3\}$ for 10-class datasets). 
% During SFL, a client in \proposal quantizes features into 16-bit before transmission to halve the communication cost.
We conduct up to $R\in\{50,120,400,800,1000\}$ SFL rounds with full participation ($|\mathbb{P}_r|=N$) unless stated otherwise, where $\mathbb{P}_r$ is the participating-client set at round $r$. Each client trains for \{1,5\} local epoch(s) per round with $\gamma=0.5$ using cross-entropy loss $\mathcal{L}(\cdot)$. 
For \proposal, we use SGD with learning rate $\beta\in\{0.025,0.05\}$, momentum 0.9, and weight decay \{0,1,5\}${\times}10^{-4}$; meta-adaptation uses inner-loop learning rate $\alpha\in\{0.01,0.05\}$ with $M\in\{3,5,10\}$ steps.
We use batch size \{50,64,128,256,512\} for training and 1 for testing. 
When reporting upper-bound performance, each client selects $e_n$ by oracle search (e.g., $e_n\in\{0.05,0.1,0.2,0.4,0.8,1.2,1.6,2.3\}$ for 10-class datasets). 
During SFL, \proposal quantizes transmitted features to 16-bit to halve communication cost.
We implement all methods in PyTorch and report 95\% confidence intervals over 3 random seeds.

% We implement our code using PyTorch. 
% The entire experiments are held over multiple GPU servers containing NVIDIA TITAN RTX, GeForce RTX 2080 Ti, 3090 Ti, A100, and GH200. 
%We avoid running experiments with the same random seed on machines whose results are not reproducible with each other. 
% We report 95\% confidence interval over 3 trials with different random seeds. 

\begin{table}
\centering
\small
\setlength{\tabcolsep}{2pt}
\begin{tabular}{llcc}
\toprule
\textbf{Scheme} & \textbf{Metric} & $\rho{=}0$ & 97/300 \\
\midrule

\multirow{2}{*}{SplitGP ($\lambda{=}0.9$)}
& Accuracy & 83.06 $\pm$ 2.79 & 62.76 $\pm$ 2.10 \\
& Latency  & 37.40 $\pm$ 3.29 & 49.54 $\pm$ 4.30 \\
\midrule

\multirow{2}{*}{SplitGP ($\lambda{=}0.2$)}
& Accuracy & 70.62 $\pm$ 3.19 & 53.74 $\pm$ 2.43 \\
& Latency  & 33.65 $\pm$ 3.88 & 44.72 $\pm$ 4.98 \\
\midrule

\multirow{2}{*}{SplitGP ($\lambda{=}0.2$) + FT}
& Accuracy & 70.73 $\pm$ 2.29 & 53.55 $\pm$ 1.76 \\
& Latency  & 38.12 $\pm$ 1.33 & 50.71 $\pm$ 1.97 \\
\midrule

\multirow{2}{*}{\textbf{HARMONY}}
& Accuracy & \textbf{83.52 $\pm$ 2.28} & \textbf{63.15 $\pm$ 1.78} \\
& Latency  & 34.26 $\pm$ 1.93 & 45.53 $\pm$ 2.56 \\
\bottomrule
\end{tabular}
\caption{Comparison of test accuracy [\%] and latency [${\times}10^{8}$] in CIFAR-100 / $N=30$ / Shard-3 / ViT-Small-16 settings.}
\label{table:main_results3}
\end{table}

\begin{table}[t]
\centering
\small
\setlength{\tabcolsep}{3pt}
\renewcommand{\arraystretch}{0.85}
\begin{tabular}{@{} c c l c c c @{}}
\toprule
Setting & ${\alpha}_{D}$ & Scheme & \makecell[c]{Max. \\ Latency} &
\multicolumn{2}{c}{Test Accuracy}\\
\cmidrule(lr){5-6}
 & & & & w/o Drift & w/ Drift \\
\midrule
% ---------- AA ----------
\multirow{12}{*}{\makecell{\scriptsize FMNIST / \\ \scriptsize $N=50$ / \\ \scriptsize 8-layer \\ \scriptsize CNN}}
 & \multirow{6}{*}{0.1}
   & SplitFed            & 6.14 & 84.43{$\pm$}1.96 & 82.78{$\pm$}2.96\\
 & & Multi-exit          & 3.18 & 84.24{$\pm$}1.52 & 80.30{$\pm$}2.60\\
 & & $\lambda{=}0.9$      & 4.12 & 94.37{$\pm$}0.81 & 56.15{$\pm$}7.61\\
 & & $\lambda{=}0.2$      & 2.36 & 90.91{$\pm$}1.27 & 78.45{$\pm$}2.47\\
 & & $\lambda{=}0.2{+}$FT & 4.34 & 93.91{$\pm$}1.37 & 75.44{$\pm$}2.44\\
 & & \proposal            & 3.31 & \textbf{94.82{$\pm$}0.60} & \textbf{85.24{$\pm$}0.60}\\
\cmidrule(lr){2-6}
 & \multirow{6}{*}{0.3}
   & SplitFed            & 6.26 & 86.75{$\pm$}2.02 & 85.92{$\pm$}1.12\\
 & & Multi-exit          & 1.74 & 84.64{$\pm$}0.80 & 83.56{$\pm$}0.69\\
 & & $\lambda{=}0.9$      & 4.20 & 91.63{$\pm$}1.81 & 69.77{$\pm$}2.55\\
 & & $\lambda{=}0.2$      & 1.77 & 89.63{$\pm$}0.95 & 82.13{$\pm$}0.09\\
 & & $\lambda{=}0.2{+}$FT & 3.38 & 92.03{$\pm$}1.51 & 79.68{$\pm$}0.22\\
 & & \proposal            & 2.84 & \textbf{93.01{$\pm$}0.67} & \textbf{87.19{$\pm$}0.36}\\
\midrule
% ---------- BB ----------
\multirow{12}{*}{\makecell{\scriptsize CINIC-10 / \\ \scriptsize $N=100$ / \\ \scriptsize VGG-11}}
 & \multirow{6}{*}{0.1}
   & SplitFed            & 38.96 & 40.61{$\pm$}13.3 & 39.59{$\pm$}11.9\\
 & & Multi-exit          & 36.80 & 47.19{$\pm$}2.69 & 29.59{$\pm$}8.15\\
 & & $\lambda{=}0.9$      & 36.67 & 72.22{$\pm$}2.70 & 26.80{$\pm$}2.90\\
 & & $\lambda{=}0.2$      & 35.84 & 59.15{$\pm$}1.69 & 31.73{$\pm$}7.48\\
 & & $\lambda{=}0.2{+}$FT & 37.29 & 70.30{$\pm$}2.47 & 31.39{$\pm$}7.97\\
 & & \proposal            & 32.60 & \textbf{81.38{$\pm$}0.75} & \textbf{47.63{$\pm$}4.10}\\
\cmidrule(lr){2-6}
 & \multirow{6}{*}{0.3}
   & SplitFed            & 38.57 & 49.18{$\pm$}12.5 & 47.77{$\pm$}11.0\\
 & & Multi-exit          & 37.85 & 46.30{$\pm$}3.64 & 35.17{$\pm$}2.63\\
 & & $\lambda{=}0.9$      & 38.42 & 63.43{$\pm$}2.41 & 32.04{$\pm$}1.29\\
 & & $\lambda{=}0.2$      & 37.73 & 53.03{$\pm$}2.77 & 37.97{$\pm$}1.04\\
 & & $\lambda{=}0.2{+}$FT & 38.21 & 61.29{$\pm$}2.02 & 37.79{$\pm$}1.61\\
 & & \proposal            & 32.10 & \textbf{75.34{$\pm$}1.90} & \textbf{52.72{$\pm$}1.75}\\
\midrule
% ---------- CC ----------
\multirow{8}{*}{\makecell{\scriptsize Tiny- \\ \scriptsize ImageNet / \\ \scriptsize $N=30$ / \\ \scriptsize ResNet-50}}
 & \multirow{4}{*}{0.3}
   & $\lambda{=}0.9$      & 39.14 & 26.81{$\pm$}0.78 & 10.72{$\pm$}4.53 \\
 & & $\lambda{=}0.2$      & 53.17 & 28.48{$\pm$}4.28 & 21.91{$\pm$}4.98 \\
 & & $\lambda{=}0.2{+}$FT & 55.91 & 29.26{$\pm$}3.35 & 20.30{$\pm$}5.03 \\
 & & \proposal            & 45.15 & \textbf{29.30{$\pm$}2.48} & \textbf{22.98{$\pm$}5.34} \\
\cmidrule(lr){2-6}
 & \multirow{4}{*}{0.5}
   & $\lambda{=}0.9$      & 28.58 & 21.29{$\pm$}0.81 & 9.55{$\pm$}1.25 \\
 & & $\lambda{=}0.2$      & 33.94 & 21.99{$\pm$}1.72 & 16.43{$\pm$}2.45 \\
 & & $\lambda{=}0.2{+}$FT & 41.64 & 21.27{$\pm$}1.15 & 13.77{$\pm$}1.24 \\
 & & \proposal            & 35.36 & \textbf{24.32{$\pm$}1.41} & \textbf{19.39{$\pm$}1.36} \\
\bottomrule
\end{tabular}
\caption{Comparison of test accuracy [\%] and maximum latency [${\times}10^{8}$] in various Dirichlet-based non-IID and drift settings. For CINIC-10, we set  $|\mathbb{P}_{r}|=10$ for each round $r$.}
\label{table:main_results4}
\end{table}

\noindent
\textbf{Metrics.} We evaluate the test accuracy and inference latency modeled in~\eqref{eq:latency_modeling}~\cite{HanKCNMCB:24}. $|\cdot|$ is the number of elements, Pr(Tx) is the offloading probability, $\text{dim}_{\text{cut}}$ is the channel dimension of cut-layer, $P_{\star}$ is computation power, and $BW$ is the uplink bandwidth. We set different $P_{C}$ across clients such that clients with smaller model assignments operate under lower power budgets. $BW=1$ and $P_{S}=100$.
\begin{equation}\label{eq:latency_modeling}
\text{Latency}=|\mathcal{D}|\left\{\frac{|\phi|\,+\,|h|}{P_{C}}\,+\text{Pr}(\text{Tx})\cdot\left(\,\frac{\text{dim}_{\text{cut}}}{BW}\,+\,\frac{|\theta|}{P_{S}}\right)\right\}
\end{equation}

\noindent
\textbf{Baselines.} We adopt representative SFL baselines below:
\begin{itemize}
    \item SplitFed (v1): trains both sides to be generalized. It doesn't support early-exit at inference.
    \item Multi-exit SFL: generalizes both sides. Equivalent to SplitFed with multi-exit or SplitGP with $\lambda=0$.
    \item SplitGP ($\lambda=0.9$): hybrid SFL in max personalization.
    \item SplitGP ($\lambda=0.2$): default SplitGP with generalization.
    \item SplitGP ($\lambda=0.2$) + fine-tuning (FT): a strategy to enhance server generalization first, then complement client personalization via on-device fine-tuning. The FT learning rate $\alpha$ and steps $M$ are same as \proposal uses.
\end{itemize}

\noindent
\textbf{Main Results.}
Tables \ref{table:main_results1}--\ref{table:main_results3} compare the overall performance of \proposal with baselines under various non-IID/drift settings. It is noteworthy that merely reducing latency without any accuracy improvement undermines the purpose of hybrid SFL and is therefore undesirable. Multi-exit SFL can reduce latency but shows OOD-tolerant yet low-level accuracy. SplitGP suffers from either severe degradation on OOD ($\lambda=0.9$) or overall sub-optimality ($\lambda=0.2$). Additional on-device fine-tuning commonly improves the ID accuracy but impairs the counterpart in most cases. In contrast, \proposal preserves the offload-when-necessary doctrine and lifts performance on both ends by up to 43.0\%/28.3\% without/with OOD, respectively, demonstrating the possibility of coexistence between strong personalization and tolerable generalization. Through Table \ref{table:main_results4}, we confirm that \proposal outperforms the baselines even under Dirichlet-based non-IID and drift conditions. \proposal achieves up to 22.0\%/15.9\% accuracy improvements over SplitGP~($\lambda$ = 0.2) without/with OOD, respectively. The gap is particularly pronounced in the CINIC-10 experiments, which can be attributed to catastrophic forgetting compounded by only 10\% of client participation probability ($N=100$, ${\forall}r\:\:|\mathbb{P}_{r}|=10$).

\begin{figure}[t]
\centerline{\includegraphics[width=0.48\textwidth]{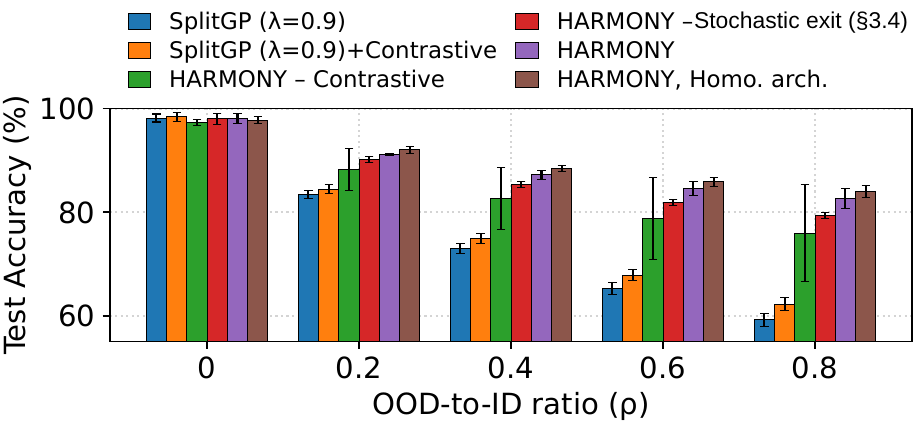}}
\caption{Ablation study on \proposal, in FMNIST / $N=50$ / Shard-2 / 8-layer CNN settings.} 
\label{fig:ablation}
\end{figure}

\noindent
\textbf{Ablation Studies.} To more closely examine the effectiveness of each module, we compare SplitGP with strong personalization, SplitGP aided by contrastive learning, \proposal without contrastive learning, \proposal with architectural heterogeneity-agnostic contrastive learning, full \proposal, and \proposal where every client architecture is fixed at the largest one, in terms of test accuracy. As shown in Figure \ref{fig:ablation}, even when contrastive learning is added to SplitGP (by introducing one classification and two contrastive branches), we observe that its OOD performance does not improve significantly. This validates that simulating fast and variant adaptation at each client through meta-learning is a crucial component for effective contrastive learning. 
% Conversely, removing contrastive learning from \proposal leads to the worse personalization–generalization trade-off observed as in SplitGP, confirming that contrastive learning is indeed the module responsible for improving OOD performance.
Conversely, removing contrastive learning from \proposal leads to a worse personalization–generalization trade-off, as observed in SplitGP, confirming that contrastive learning is indeed the module that improves OOD performance.
Finally, our stochastic early feature extraction is proven to be effective exactly under architectural heterogeneity, since the \proposal with homogeneous client architecture shows high performance even without stochastic early feature extraction.

\begin{figure}[t]
  \centering
  \begin{subfigure}[t]{0.48\linewidth}
    \centering
    \includegraphics[width=\linewidth]{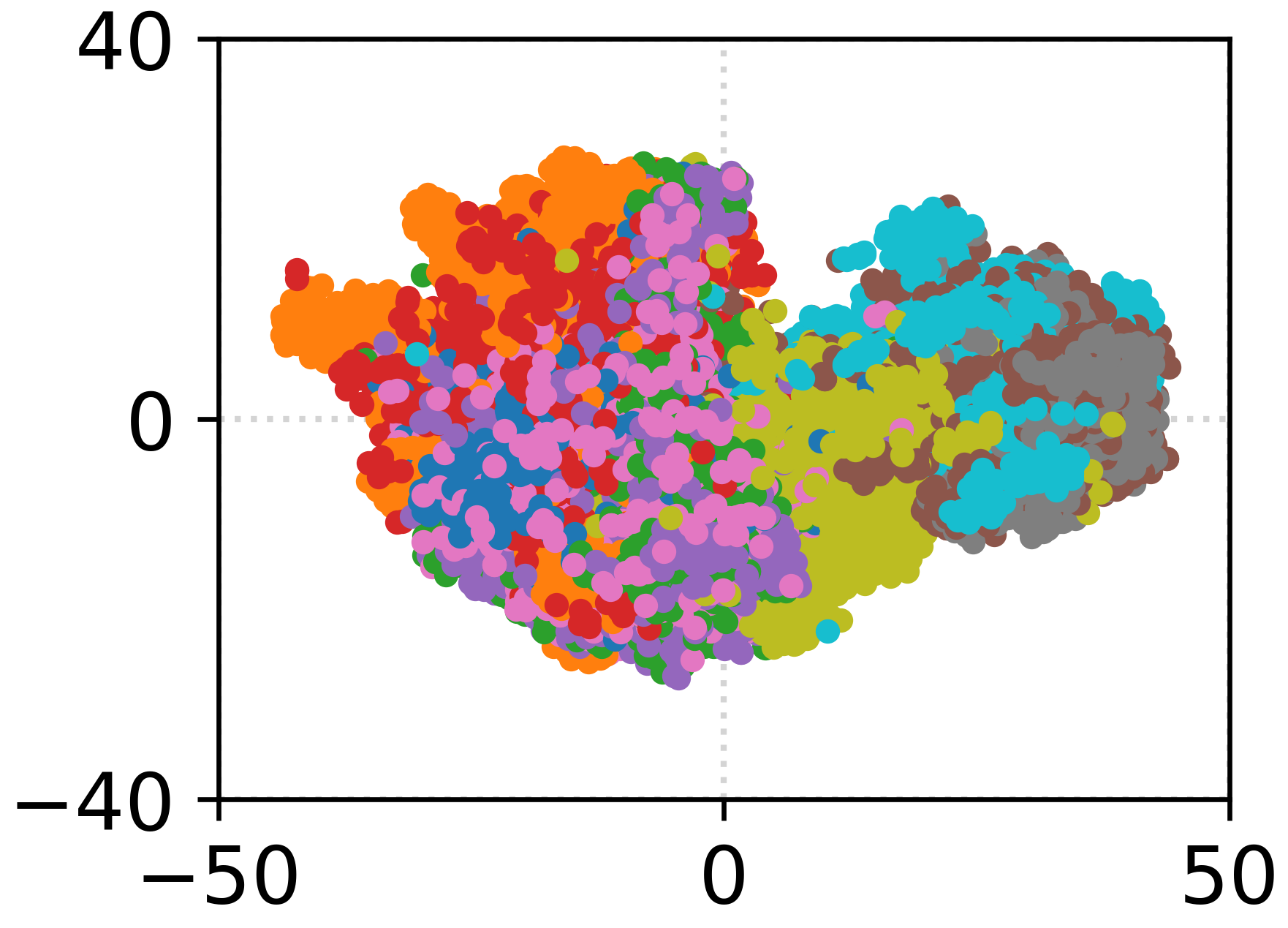}
    \caption{SplitGP ($\lambda=0.9$)}\label{fig:tsne_compared}
  \end{subfigure}
  \begin{subfigure}[t]{0.48\linewidth}
    \centering
    \includegraphics[width=\linewidth]{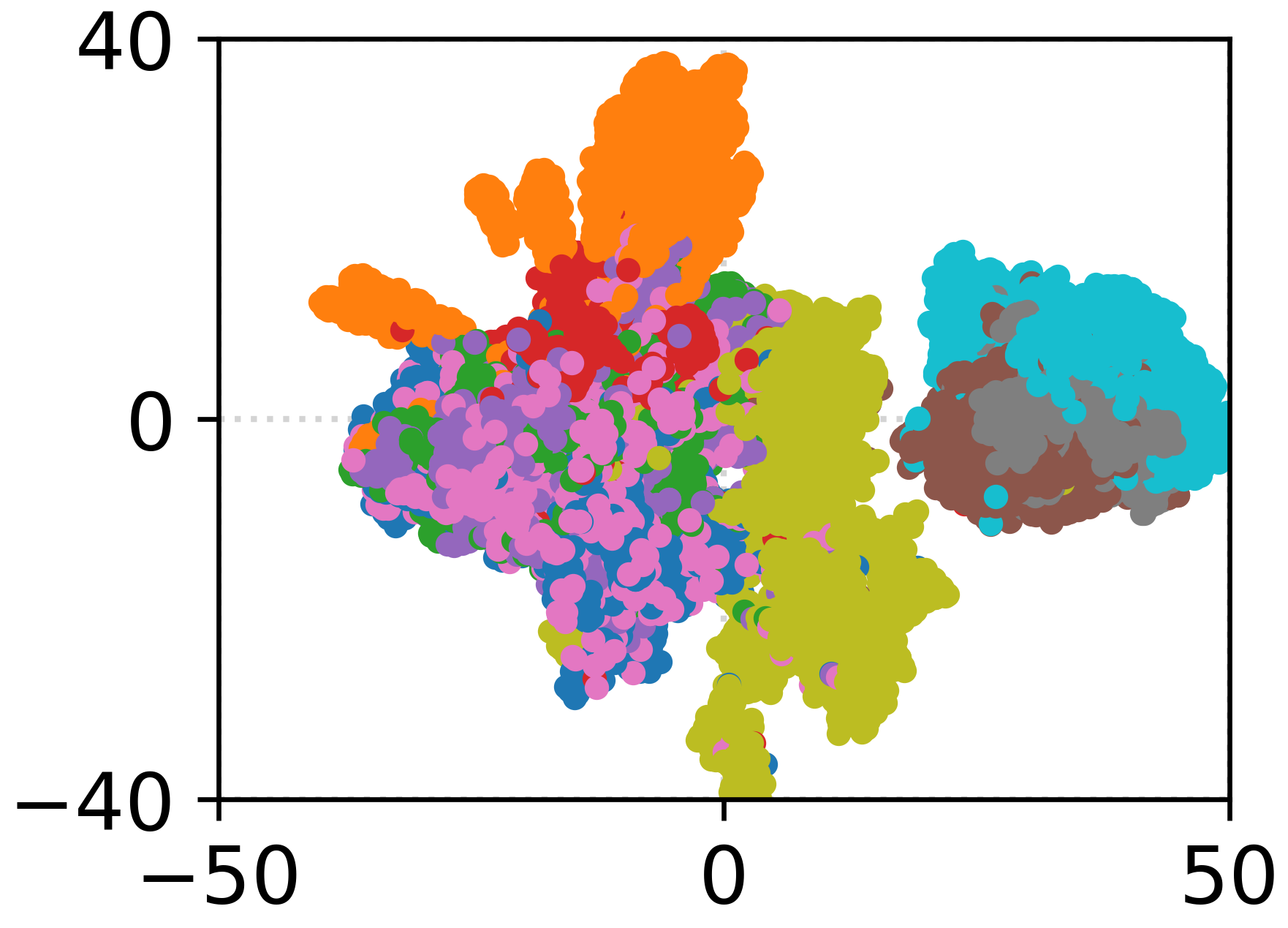}
    \caption{\proposal}\label{fig:tsne_proposed}
  \end{subfigure}
  \caption{Representation space in the main server's perspective with FMNIST / $N=50$ / Shard-2 / 8-layer CNN, expressed by T-SNE. It consists of server-side representations (right before ${\theta}_{\text{head}}$) from customized client extractors. Each color represents a data class.}\label{fig:tsne}
\end{figure}

\noindent
\textbf{Representation Space Visualization.} To empirically validate that contrastive learning drives the main server to focus on feature patterns closely tied to data classes, we visualize the representation space used for prediction at the server—specifically, the activations immediately before the classifier head in $\theta$, using T-SNE~\cite{MaatenH:2008}. Figure~\ref{fig:tsne} verifies the effectiveness of our contrastive learning methodology.

%% file: Section/6_discussion.tex
% As a pioneering effort, this paper considers heterogeneity only in the model split point across clients. In more realistic settings, however, the underlying backbone architectures of clients may fundamentally differ (e.g., CNN vs. ViT), in which case representation skew is expected to be further exacerbated. This naturally raises several open questions, including what architecture the server model $\theta$ should adopt to support all clients, how aggregation should be performed under such heterogeneity, and whether more effective alignment methodologies tailored to this setting can be developed.

% Although this paper mainly investigates class shift, hybrid SFL may exhibit additional weaknesses under other well-studied data drift types, including covariate shift~\cite{MclaughlinS:24} and domain shift~\cite{WangWWFW:25}.

% Currently, an offloading sensitivity $e_{n}$ is selected in an oracle manner to report the upper-bound performance; a practical strategy without access to the test-data labels is required.

As an initial step toward heterogeneity-aware hybrid SFL, this paper focuses on heterogeneity in the model split point across clients. In more realistic deployments, clients may adopt fundamentally different backbone architectures (e.g., CNN vs.\ ViT), where representation skew can be further exacerbated. This raises important directions for future work, including designing a server model $\theta$ that can serve multiple client families, aggregation under cross-architecture heterogeneity, and alignment methods tailored to richer architectural diversity.

While we primarily study class shift, hybrid SFL may also be influenced by other data drifts, including covariate shift~\cite{MclaughlinS:24} and domain shift~\cite{WangWWFW:25}, motivating drift-aware robustness in the hybrid offloading pipeline.

Finally, we select the offloading sensitivity $e_n$ to characterize the upper-bound tradeoff; developing a practical label-free strategy to set or adapt $e_n$ remains important for deployment.

%% file: Section/7_conclusion.tex
We propose \proposal, a heterogeneity-aware hybrid SFL framework that mitigates the personalization--generalization trade-off under model-heterogeneous settings.
By addressing representation skew from heterogeneous and strongly personalized on-device extractors, \proposal uses supervised contrastive alignment to impose class-aware structure in the shared server representation space while preserving device-specific adaptations.
Across multiple datasets, model families, and non-IID/OOD test regimes, \proposal consistently improves over state-of-the-art hybrid SFL baselines, achieving up to 43.0\%/28.3\% accuracy gains without/with OOD, respectively.
Overall, explicit alignment enables reliable server-side fallback without sacrificing on-device personalization, advancing practical federated intelligence for resource-diverse mobile ecosystems.